\documentclass[a4paper,fleqn]{cas-dc}

\usepackage[numbers]{natbib}
\usepackage[caption=false,font=normalsize,labelfont=sf,textfont=sf]{subfig}
\usepackage{subfloat}
\def\tsc#1{\csdef{#1}{\textsc{\lowercase{#1}}\xspace}}
\tsc{WGM}
\tsc{QE}

\begin{document}
\let\WriteBookmarks\relax
\def\floatpagepagefraction{1}
\def\textpagefraction{.001}

\shorttitle{}    

\shortauthors{Lintao Yang et al.}  

\title [mode = title]{Node-level Graph Neural Architecture Search Framework}  



%
\affiliation[1]{organization={College of Information Science and Engineering, Ningbo University},
            city={Ningbo},
            postcode={315211},
            country={China}}
\affiliation[2]{organization={Department of Computer Science and Technology, University of Cambridge},
            city={Cambridge},
            postcode={CB3 0FD},
            country={UK}}
\affiliation[3]{organization={School of Artificial Intelligence, Jilin University},
            city={Changchun},
            postcode={130015},
            country={China}}

\author[1]{Lintao Yang}\fnmark[1] \ead{2211100291@nbu.edu.cn}
\credit{Conceptualization; Formal analysis; Writing – original draft; Methodology; Writing – review \& editing}
\author[1]{Sirui Li}\fnmark[1] \ead{2411100282@nbu.edu.cn}
\credit{Conceptualization; Formal analysis; Writing – original draft; Methodology; Writing – review \& editing}
\author[1]{Yaqing Wang} \ead{2511100088@nbu.edu.cn}
\credit{Investigation; Validation; Writing – review \& editing}
\author[2]{Pietro Liò} \ead{Pietro.Lio@cl.cam.ac.uk}
\credit{Investigation; Software; Resources}
\author[3]{Xu Shen} \ead{196003572@nbu.edu.cn}
\credit{Methodology; Writing – original draft; Investigation}
\author[1]{Baisong Liu} \ead{liubaisong@nbu.edu.cn}
\credit{Writing – review \& editing; Supervision; Resources}
\author[1]{Chengbin Peng}\cormark[1]\ead{pengchengbin@nbu.edu.cn}
\credit{Formal analysis; Methodology; Resources; Supervision; Project administration; Writing – review \& editing}
\cortext[1]{Corresponding author}
\cormark[1] 
\fntext[1]{These authors contributed equally to this work.}


\begin{abstract}
In recent years, Graph Neural Networks (GNNs) and architecture search frameworks have gained extensive application in non-Euclidean data processing, attributable to their superior capacity in managing unstructured data. Nevertheless, traditional approaches typically apply uniform convolution operations to all nodes, regardless of their varying structural and feature characteristics, which can undermine model performance and result in over-smoothing issues as the number of layers increases. To overcome this limitation, in this work, we propose a \textbf{N}ode-Level \textbf{G}raph \textbf{N}eural \textbf{A}rchitecture \textbf{S}earch (N-GNAS) algorithm. It can automatically choose an appropriate network architecture for each subset of nodes when updating node features. N-GNAS also introduces a contrastive learning loss to separate sample features from different categories and vice versa. In experiments conducted on eight datasets for node and graph classification, our methodology outperforms current leading GNAS techniques and traditional human-designed GNNs. For example, it achieves an accuracy rate of 78.26\% on the CiteSeer dataset.
\end{abstract}





\maketitle

\section{Introduction}
Graphs are an essential representation of data structure that can describe various relationships and structures, which has been a pivotal area of investigation for many years. Graph neural networks (GNNs) \cite{battaglia2018relational, kipf2016semi} are a kind of neural network model specially designed to deal with graph structure data. They extract features of graph data by learning relationships from nodes and edges. They can perform many graph learning tasks, such as node classification \cite{wang2019heterogeneous}, link prediction \cite{kumar2020link, liben2003link, lu2011link}, and graph classification \cite{xu2018powerful}. 

Despite the success of GNNs \cite{li2023graph, wang2023tcam, shen2024optimizing, miao2024rethinking, chen2023sp, xue2024psa}, huge challenges remain. For example, a specific graph neural architecture can usually only fit some graph learning tasks on certain datasets, and even for these datasets, lots of human expertise and trials are typically needed to optimize the graph neural architecture. Thus, it undoubtedly requires significant computational investments and professional skills.

Inspired by Neural Architecture Search (NAS) on CNNs \cite{zoph2016neural}, Graph Neural Architecture Search (GNAS) \cite{gao2023commgnas} techniques have been further proposed to search for the best GNN automatically. For example, in the beginning, GraphNAS \cite{gao2021graph} pioneered this research field by using reinforcement learning to design graph neural architectures and proposing a novel weight-sharing strategy to reduce the number of parameters to effectively select graph neural network components such as attention functions and attention heads. For heterogeneous graphs, HGNAS \cite {gao2021heterogeneous} designs a new search space by utilizing popular heterogeneous graph neural networks such as RGCN \cite {schlichtkrull2018modeling} and MAGNN \cite{fu2020magnn}, and proposes different kinds of message encoding and aggregation methods. SANE \cite{huan2021search} introduces a differentiable gradient graph neural architecture search method \cite{liu2018darts} to explore feature aggregation strategies. ARGNP \cite{cai2022automatic} introduces a relation-aware message passing strategy into the search space, to optimize the learning of node and relational information.

\begin{figure*}
	\centering
	\includegraphics[width=140mm]{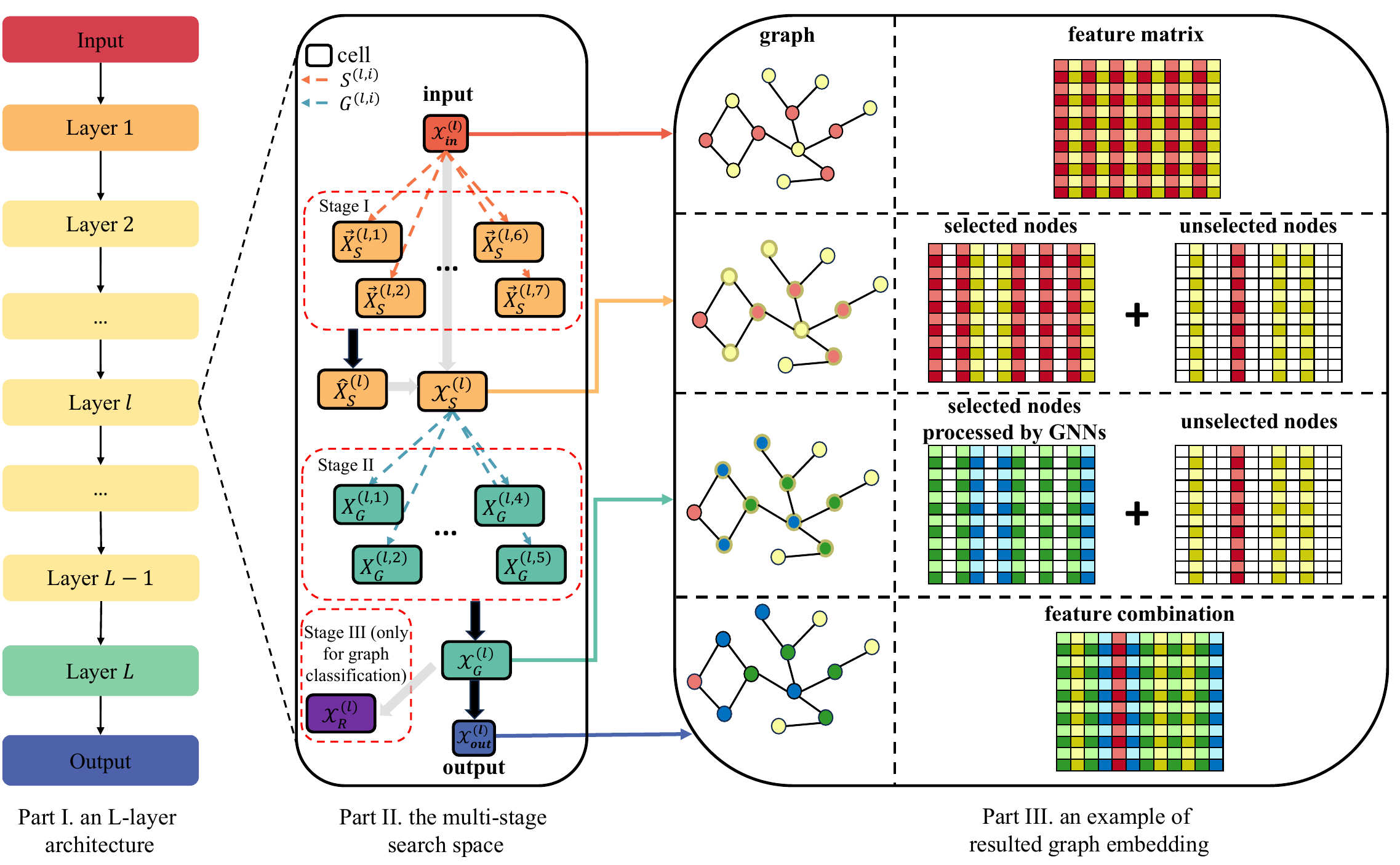}
	\caption{\textbf{An example of a graph neural network discovered by N-GNAS.} Part I illustrates an $L$-layer architecture. Part II demonstrates the search space across three stages, and Part III provides an example of search outcomes. Each layer comprises two stages: node selection operation search and graph neural network operation search, denoted by the symbols $S$ and $G$ with subscripts reflecting their diversity. Every layer comprises an input cell, an output cell, and operation cells. In this example, seven cells and five operation are used in the first and the second stage respectively. The variable $X$ denotes the respective features obtained from these cells, with subscripts distinguishing stages and superscripts defining layers and cells. Links indicate possible operations.}
	\label{img:1}
\end{figure*}

However, traditional GNAS methods usually perform a uniform operation over different nodes, which is inappropriate as nodes in a real-world graph can usually encompass various types with diverse characteristics. Meanwhile, performing the same number of graph neural network operations on each node may result in excessive averaging of node features across the entire graph, reducing differentiation between different nodes and causing the problem of over-smoothing \cite{shen2024graph}. To address this challenge, in this work, we propose a node level Graph Neural Architecture Search (N-GNAS) framework, which can adaptively apply different GNN operations for different nodes and alleviate over-smoothing. Figure \ref{img:1} illustrates the graph neural network model that has been generated using the N-GNAS algorithm.

This study represents the first effort to apply GNAS to optimize node-level operations within graph neural networks. The primary contributions of this research are outlined as follows:
\begin{enumerate}
	\item We propose a node-level search framework that can automatically select the optimal graph neural network (GNN) architecture for different node subsets based on their unique characteristics. Unlike traditional GNN or GNAS approaches, which apply uniform convolution operations to all nodes, this is the first to achieve node-level optimization for more precise and effective learning. 
	
	\item We propose a contrastive learning loss into GNAS to separate sample features from different categories and vice versa, which can also effectively alleviate the over-smoothing issue commonly observed in traditional deep GNNs.
	
	\item Extensive evaluations across diverse graph learning tasks have demonstrated that our approach surpasses conventional GNAS and manually crafted GNN methods in performance.
\end{enumerate}

\section{Related Work}
\subsection{Recent Advances In Graph Neural Networks}
Graph Neural Networks (GNNs) is a type of deep learning model used for processing graph-structured data \cite{kipf2016semi, hamilton2017inductive, velivckovic2017graph, xu2018powerful} that can learn representations for nodes and graphs and can perform various tasks such as node classification, graph classification, link prediction, and so on. GNN operations can aggregate node information, and GNNs can be divided into two categories: spectral-based GNNs \cite{kipf2016semi} and spatial-based GNNs \cite{velivckovic2017graph}.

Spectral-based GNNs learn node representations by spectral decomposition of graphs. Specifically, for a graph, the eigen-decomposition of its Laplacian matrix is computed, and these eigenvectors are then used as node representations. Defferrard et al. \cite{defferrard2016convolutional} approximate spectral filters using Chebyshev polynomials to reduce the complexity of matrix decomposition. GCN \cite{kipf2016semi} optimized upon this by simplifying the previous method through local first-order approximations. JK-Net \cite{xu2018representation} improves node representations by adaptively aggregating nodes from different locations. SGC \cite{wu2019simplifying} simplifies GCN by removing non-linear activation functions and feature extraction as a feature preprocessing operation.

Spatial-based GNNs directly utilize the neighborhood information to learn node representations, which update node features simply by aggregating information within the local neighborhood of each node and do not require expensive graph spectral decomposition. For example,  GraphSAGE \cite{hamilton2017inductive} is a framework that leverages both node feature information and node structural information. The feature information is sampled and aggregated from the local neighborhood of nodes to obtain node embeddings. The structural information is learned by employing trainable aggregation functions. GAT \cite{velivckovic2017graph} introduces a masked self-attention mechanism to assign different weights for different node representations. GatedGCN \cite{bresson2017residual} replaces the recursive definition of GNNs with gated edges and residuality, employing gate mechanisms to determine whether information should propagate from one node to another and utilizing residual connections to enable the network to learn features of graph data at deeper levels.

\subsection{Graph Neural Architecture Search}
The objective of NAS is to autonomously identify the most effective neural architecture tailored for specific datasets through a systematic optimization process \cite{zoph2016neural}. Traditionally, the design of neural network architectures is conducted manually, a process that is not only laborious but also time-consuming \cite{liu2018darts}. To overcome this limitation, NAS can automatically discover suitable network architectures for predefined tasks. Similar to NAS on CNNs and RNNs, NAS on graph has been proposed \cite{huan2021search, cai2022automatic}.
For example, there are GNAS based on reinforcement learning (RL) \cite{zoph2016neural,gao2021graph,zhao2020simplifying}, GNAS based on evolutionary algorithms (EA) \cite{liu2017hierarchical,shi2022genetic,knowledge-aware,zhang2022deep}, GNAS based on gradient descent (GD) \cite{liu2018darts,huan2021search,d2gnas,wei2023neural}, etc. These algorithms can not only automatically search the network structure but also avoid the "subjective" bias of the designer for the network structure.

In search space design, prevailing approaches are generally categorized into two main paradigms: component-based and architecture-based methods \cite{schlichtkrull2018modeling}. The component-based search method targets a predefined set of components, such as aggregation operations, their respective weights, and combination mechanisms. It aims to optimize each module by fine-tuning its attributes and configurations. In contrast, architecture-based search involves a broader exploration scope, including elements like network depth and integrating different modules, to identify the optimal configuration for the entire network's structure \cite{zhang2021automated}. Futhermore, DSS \cite{one_shot_gnas} further improves search efficiency by adaptively pruning the search space during the search process.

Nevertheless, these approaches overlook the diversity of node attributes and solely rely on cross-entropy as the loss function. Unlike heterogeneous GNN methods \cite{schlichtkrull2018modeling, fu2020magnn, gao2021heterogeneous} that rely on {predefined} type information and manual metapaths, our approach {automatically} determines node-level operations from learned features and structures without requiring any type annotations, making it applicable to both homogeneous and heterogeneous graphs in a fully data-driven manner. Based on these insights, we propose an approach for the node-level graph neural architecture search.

\section{METHODOLOGY}
\subsection{Notations}
{
In this work, we define $G$ as a graph of $n$ nodes, with $V$ representing the set of nodes within $G$. We denote the node features by $\mathcal{X} \in {\mathbb R}^{n \times {d}_{V}}$, where ${d}_{V}$ indicates the dimension of node features. Specifically, the $u$-th row of $\mathcal{X}$, corresponding to node $u$ is denoted as $x_{u}$. We use $N(v)$ to denote the one-hop neighbors of a node $v$ in $G$. We use $A \in {\mathbb R}^{n \times n}$ and $D \in {\mathbb R}^{n \times n}$ to represent the adjacency matrix and the degree matrix of this graph. Thus, the structural relationships and the node features in graph $G$ can be represented by an adjacency matrix $A$  and a feature matrix $\mathcal{X}$.
}
\subsection{Search Space Architecture Design}
Traditional GNAS applies the same number of graph neural networks to all nodes in the graph, which cannot adapt well to the diversified characteristics of different nodes, thus limiting the expressive power of the graph neural network model. In this work, we propose applying varying numbers of graph neural networks to different nodes based on their characteristics and needs. For example, by considering node features, positions, and local structures, we tailor the number of graph neural networks to enhance the model's expressive capacity.

Without loss of generality, we assume that the network architecture consists of $L$ layers. Each layer consists of node selection operations and graph neural network operations, except that Layer 1 and Layer $L$ are usually linear and classification layers, respectively. We use $\mathcal{X}_{in}^{(l)}$ and $\mathcal{X}_{out}^{(l)}$ to represent the input and the output of the $l$-th layer respectively. Thus, $\mathcal{X}_{in}^{(1)}$ is the input graph and $\mathcal{X}_{out}^{(L)}$ is the network output, and $\mathcal{X}^{(l)}_{in}=\mathcal{X}^{(l-1)}_{out}$, for $l\in \left \{ 2,3,\dots,L \right \} $. 

Each layer encompasses two stages. The first is to select nodes for further processing, and the second is to search for optimal graph neural network operations for selected nodes. Hence, the search space can be represented by a directed acyclic graph (DAG). Cells are defined as the fundamental elements within the search space, and links represent the connections between these cells. For the $l$-th layer, a vector $\vec{X}_S^{(l, i)} \in {\mathbb R}^{1 \times n}$ to represents the node information score in the $i$-th cells corresponding to the $i$-th scoring strategy in the first stage, and a matrix $\mathcal{X}_G^{(l, i)} \in {\mathbb R}^{n \times n}$ represents the feature embedding corresponding to the $i$-th graph neural network operation in the second stage. There is a total of $n_1$ and $n_2$ cells for corresponding stages. In each stage, an operation depicted by a directed link $S^{(l, i)}$ or $G^{(l, i)}$ extracts features from its input cell and stores the results in its output cell. The framework is as shown in Figure \ref{img:1}.

\subsubsection{Stage I: Node Selection Operations.} Before applying a graph neural network operation in each layer, we propose different scoring approaches and ensemble them to compute a soft gating vector, where each entry indicates the extent to which a node should undergo further GNN processing. Without loss of generality, for the $l$-th layer and the $i$-th scoring approach, node information score $\vec{X}_S^{(l,i)}$ are obtained after applying an operation $S^{(l,i)}$ on the input cell $\mathcal{X}_{in}^{(l)}$. After assembling these scoring approaches, nodes with higher scores are assigned larger gate values, allowing them to contribute more to the subsequent GNN stage, while nodes with lower scores are mostly routed through residual connections. 

Node scoring approaches consider node features, positions, and structures, and we discuss each in detail as follows.

\textbf{ (1) Feature Projection $f_{S_{1}}$}: 
From the perspective of node features, we calculate the score for each node by employing a learnable feature projection vector $W^{(l)}$ as follows: 
\begin{align}
	\label{XS_F1}
	f_{S_{1}}(\mathcal{X}_{in}^{(l)}) &= W^{(l)} \cdot \mathcal{X}_{in}^{(l)}.
\end{align}

\textbf{ (2) Multi-layer Perceptron $f_{S_{2}}$}: We also employ a Multi-Layer Perceptron ($MLP$) to obtain a score vector for the entire feature matrix for the score calculation as follows:
\begin{align}
	\label{XS_F2}
	f_{S_{2}}(\mathcal{X}_{in}^{(l)}) &= MLP 
	(\mathcal{X}_{in}^{(l)}). 
\end{align}

\textbf{ (3) Local Position Score $f_{S_{3}}$}:
Regarding node positions, we can consider local and global aspects. The local position score is computed by generating a random walk matrix that quantifies the likelihood of each node reaching other nodes within the graph. The score of each node is obtained by summing each column of the random walk matrix: 

\begin{align}
	\label{XS_P3}
	M &= D \cdot A, \\
	f_{S_{3}}(\mathcal{X}_{in}^{(l)}) &= \left [ \sum_{j}^{n}M_{j,1}, \sum_{j}^{n}M_{j,2}, \cdots, \sum_{j}^{n}M_{j,n} \right ].
\end{align}

\textbf{(4) Global Position Score $f_{S_{4}}$}: The global position score is derived using the distance matrix $P \in {\mathbb R}^{n \times n}$ between nodes, in which each column indicates the distance from each node to every other node and can be computed by the Floyd–Warshall \cite{floyd1962algorithm} algorithm or approximated by more efficient methods for large graphs. The score for each node is obtained by summing elements of each column in the matrix as follows:

\begin{align}
	\label{XS_P4}
	f_{S_{4}}(\mathcal{X}_{in}^{(l)}) &= \left [ \sum_{j}^{n}P_{j,1}, \sum_{j}^{n}P_{j,2}, \cdots, \sum_{j}^{n}P_{j,n} \right ].
\end{align}

\textbf{(5) Local Structural Feature Score $f_{S_{5}}$}: We also use node structural features for scoring. Local structure information is directly obtained from the degree matrix as follows: 

\begin{align}
	\label{XS_S5}
	f_{S_{5}}(\mathcal{X}_{in}^{(l)}) = \left [ D_{1,1}, D_{2,2}, \cdots, D_{n,n} \right ].
\end{align}

\textbf{ (6) Relative Structural Feature Score $f_{S_{6}}$}: Relative structural feature scores leverage node features from neighborhoods and are obtained by aggregating their feature projection scores  $ f_{S_{1}}(\mathcal{X}_{in}^{(l)})$, rather than simply the number of neighbors as follows: 

\begin{align}
	\label{XS_S6}
	\psi  &= f_{S_{1}}(\mathcal{X}_{in}^{(l)}), \\
	f_{S_{6}}(\mathcal{X}_{in}^{(l)}) &= \left [ \sum_{j \in N(v_{1})}\psi_{j}, \sum_{j \in N(v_{2})}\psi_{j}, \cdots, \sum_{j \in N(v_{n})}\psi_{j} \right ].
\end{align}

\textbf{ (7) Attention-based Structural Feature  Score $f_{S_{7}}$}: Adaptive Structure Aware Pooling ($ASAP$) \cite{ranjan2020asap} can compute node attentions to divide nodes in a graph into different local clusters. Thus, we propose to use this attention information as node scores as follows:

\begin{align}
	\label{XS_S7}
	f_{S_{7}}(\mathcal{X}_{in}^{(l)}) = {ASAP}(\mathcal{X}_{in}^{(l)}).
\end{align}

We use $\vec{X}_{S}^{(l,i)}$ to represent the scores obtained from these scoring approaches:

\begin{align}
	\label{XS}
	\vec{X}_{S}^{(l,i)} &= Norm(f_{S_{i}}(\mathcal{X}_{in}^{(l)})),\ i \in \left \{ 1,2,\dots,7 
	\right \}.
\end{align}	

An activation function $\sigma$ is applied on the ensemble of these scores to compute the node mask vector $\hat{X}_{S}^{(l)} \in {\mathbb R}^{1 \times n}$ of the $l$-th layer based on that from the last layer, namely, $\hat{X}_{S}^{(l-1)}$: 

\begin{align}
	\label{XS1}
	\hat{X}_{S}^{(l)}  &= \sigma (\sum_i \alpha_{S}^{(l,i)} \vec{X}_{S}^{(l,i)}) \odot \hat{X}_{S}^{(l-1)},   
\end{align}	
where $\alpha_{S}^{(l,i)}$ is a scalar representing the weight of the $i$-th operation, $\odot$ is an element-wise multiplication operation, and $\sigma(\cdot)$ is the sigmoid function ensuring each entry lies in $(0,1)$. Thus, $\hat{X}_{S}^{(l)} \in (0,1)^n$ serves as a continuous soft gating vector of the $l$-th layer, where the $j$-th entry indicates the degree to which node $j$ should be processed by subsequent GNN operations. A value close to $1$ means the node is strongly selected, while a value close to $0$ means its features mostly pass through via the residual connection in Eq.~\eqref{XG}. This continuous relaxation keeps the entire framework fully differentiable without requiring any hard discretization during training. 

After obtaining $\hat{X}_{S}^{(l)}$, we zero out the corresponding columns in the feature matrix $\mathcal{X}_{in}^{(l)}$ row by row to obtain $\mathcal{X}_S^{(l)}$ for further processing:
\begin{align}
	\label{S}
	\mathcal{X}_S^{(l)}= \hat{X}_{S}^{(l)}\odot \mathcal{X}_{in}^{(l)},
\end{align}
where broadcasting is used in $\odot$ to extend a vector to match the dimensions required for matrix multiplication.

\subsubsection{Stage II: Graph Neural Network Operations.} 

In the second stage, each cell also connects with the cell $\mathcal{X}_S^{(l)}$ and stores the results obtained from different graph neural network operations. We can select the optimal graph neural network operation for the current layer by analyzing the final results, thus gaining advantages in graph neural architecture search.

Different graph neural network operations have distinct advantages over different tasks and datasets. In this work, several widely used GNNs are included in the search space to enhance the model performance, as shown in Table \ref{tab:1}, where each operation $f_{G_{i}}$ is fed by a graph and the corresponding feature matrix, and returns the resulted feature matrix. 

\begin{table}[htp]
	\caption{Some popular graph neural network operation within our search space. \textbf{Def.} denotes different graph neural network operations. \textbf{Operation} denotes the corresponding concrete graph neural network operation respectively.}
	\label{tab:1}       
	\centering
	\resizebox{0.2\textwidth}{0.7in}{
		\begin{tabular}{c | c }     
			\hline\noalign{\smallskip}       
			\textbf{Def. } & \textbf{Operation} \\   
			\noalign{\smallskip}\hline\noalign{\smallskip}
			
			\noalign{\smallskip}
			$f_{G_{1}}$ &
			GCN \cite{kipf2016semi}  \\ 
			\noalign{\smallskip}\hline\noalign{\smallskip}
			$f_{G_{2}}$ &
			GAT \cite{velivckovic2017graph}\\
			\noalign{\smallskip}\hline\noalign{\smallskip}
			$f_{G_{3}}$ &
			GraphSAGE \cite{hamilton2017inductive}\\
			\noalign{\smallskip}\hline\noalign{\smallskip}
			$f_{G_{4}}$ &
			GIN \cite{xu2018powerful} \\
			\noalign{\smallskip}\hline\noalign{\smallskip}
			$f_{G_{5}}$ &
			GatedGCN \cite{bresson2017residual} \\
			\noalign{\smallskip}\hline
	\end{tabular}}
	
\end{table}

Therefore, the objective function can be expressed as follows:

\begin{align}
	\label{XG}
	X_{G}^{(l,i)} = f_{G_{i}}(\mathcal{X}_{S}^{(l)}) + (1 - \hat{X}_{S}^{(l)}) \cdot \mathcal{X}_{in}^{(l)},
\end{align}
where the term $(1 - \hat{X}_{S}^{(l)}) \cdot \mathcal{X}_{in}^{(l)}$ acts as a residual connection: nodes with low gate values are directly carried over without further GNN processing, which helps alleviate over-smoothing. Through node-level graph neural network operation selection, we can obtain the final node features $\mathcal{X}_G^{(l)}$ for each layer:

\begin{align}
	\label{G}
	\mathcal{X}_G^{(l)}=\sum_{1 \le i < 5}\alpha_{G}^{(l,i)}X_{G}^{(l,i)}.
\end{align}

where $\alpha_{G}^{(l,i)}$ denotes the scalar weight for the $i$-th operation.

\subsubsection{Stage III: Readout Operations For Graph Classification Tasks.}

Although for node classification tasks, node feature matrix $\mathcal{X}_G^{(l)}$ can be directly used for subsequent classification tasks, for graph classification tasks, it is necessary to perform a Readout operation \cite{ying2018hierarchical} to summarize the obtained feature matrix. In this stage, we discuss the latter case. 

\begin{figure*}[ht]
	\centering
	\includegraphics[width=140mm]{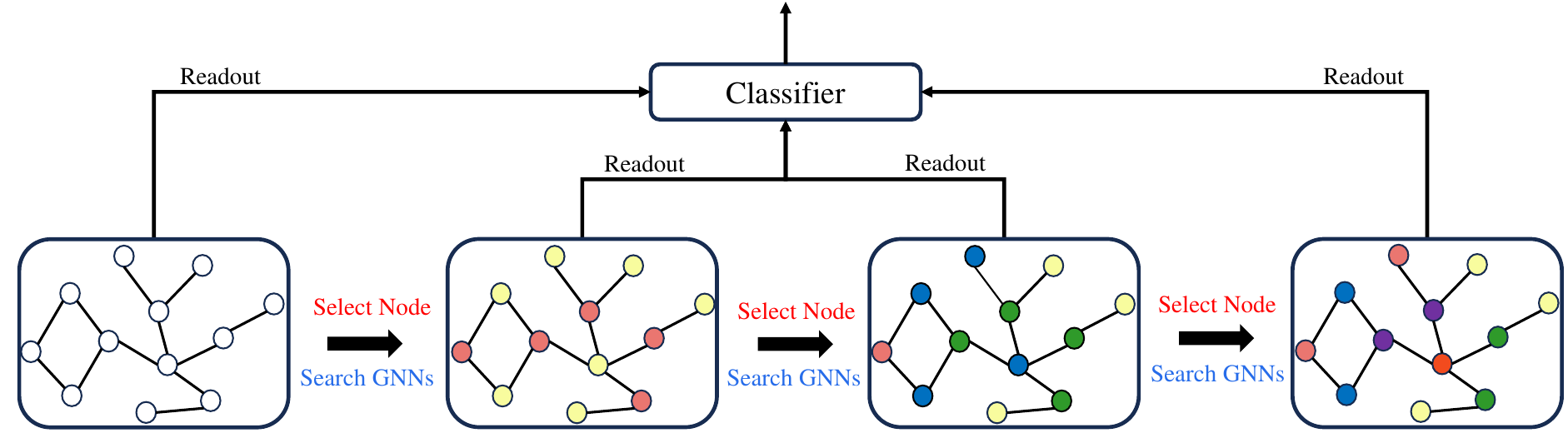}
	\caption{\textbf{The specific process of N-GNAS in graph classification task.} \textcolor{red}{Select Node} refers to the process of searching for node selection operations. In contrast, \textcolor{blue}{Search GNNs} indicates the process of searching for graph neural network operations. A \textbf{Readout} operation is required for the result of each layer, and the \textbf{Readout} result is used as the output of the current layer.}
	\label{Readout}
\end{figure*}

We include common Readout operations such as \\ GLOBAL\_MEAN, GLOBAL\_MAX, GLOBAL\_SUM, \\ GLOBAL\_SORT, GLOBAL\_ATT, and ZERO in the search space to explore the best Readout method for each layer, aiming to obtain better global graph features. An example of this procedure is depicted in Figure \ref{Readout}.

\subsection{Search Strategy Design}

In this part, we discuss the loss functions and the search strategies. 
\subsubsection{Loss Function}

The loss functions are composed of two parts: a traditional cross-entropy (CE) fit of the training data and a proposed contrastive learning (CL) Loss to balance model performance and alleviate the problem of over-smoothing.

\textbf{Cross-entropy (CE) Loss.} The metric of cross-entropy \cite{de2005tutorial} is frequently utilized in classification problems. When a model's objective is to minimize this loss, it aligns the predictions more closely with the actual labels, thereby improving its predictive performance. They commonly gauge the disparity between model predictions and actual labels. By minimizing the cross-entropy loss, the model can better fit the training data and enhance accuracy in classification tasks. The cross-entropy loss in our context is defined as follows: 
\begin{align}
	\label{Lce}
	L_{ce} = -\frac{\sum_{i=1}^{N}y_{i}\log\widehat{y}_{i} + (1-y_{i})\log(1-\widehat{y}_{i})}{N},
\end{align}
where $y_{i}$ and $\widehat{y}_{i}$ represent the ground-truth label and the predicted label of sample $i$, respectively, $N$ represents the number of nodes in node classification tasks or the number of graphs in graph classification tasks, respectively.

\textbf{Contrastive Learning (CL) Loss.}  
Recent studies highlight the importance of over-smoothing issues in model performance. Thus, besides the CE loss, we propose a contrastive learning loss. While such losses have been adapted in computer vision or GNN, they have seldom been used in GNAS. We propose contrasting the positive and negative samples with a sampling strategy to enhance node-level feature learning. 

Specifically, we randomly select $M$ samples from the training set and use $m_i$ to represent the $i$th sample, which represents the node feature of the $i$-th node in node classification tasks or the global graph feature of the $i$-th graph in graph classification tasks. We use cosine similarity to measure the distance between sample $i$ and sample $j$ as follows:
\begin{align}
	\label{Los}
	cos(m_{i},m_{j}) &= \frac{m_{i} \cdot m_{j}}{||m_{i}|| \cdot ||m_{j}||}, \\
	sim(m_{i},m_{j}) &= exp(\frac{cos(m_{i},m_{j})}{\tau}),
\end{align}
where $exp()$ is the exponential function, $\tau$ is a temperature parameter. A pair of samples with the same ground-truth label is considered a positive sample pair, and vice versa, so we can obtain the positive and the negative distances.
\begin{align}
	pos &= \sum_{i=1}^{M}\sum_{j=1}^{M}\delta_{y_{i}=y_{j}}sim(m_{i},m_{j}), \\
	neg &= \sum_{i=1}^{M}\sum_{j=1}^{M}\delta_{y_{i}\ne y_{j}}sim(m_{i},m_{j}).
\end{align}	
where $\delta_{y_{i}=y_{j}}$ equals $1$ if $y_{i}=y_{j}$, and $\delta_{y_{i}=y_{j}}$ equals $0$ otherwise. $pos$ denotes the contribution of positive samples, and $neg$ represents the contribution of negative samples. 	

Finally, the loss function is  
\begin{align}
	L_{cl} &=  -\frac{pos}{pos + neg}.
\end{align}
It enhances the similarity between positive samples and reduces the similarity between negative samples. 

\textbf{Total Loss.} Therefore, the final loss function $L$ can be written as
\begin{align}
	\label{L}
	L = L_{ce} + L_{cl}.
\end{align}

\subsubsection{Search Strategy}

Similar to numerous other GNAS methodologies, we utilize the Differentiable Architecture Search (DARTS) \cite{liu2018darts, huan2021search} to investigate the designed search space. By leveraging a continuous relaxation strategy, DARTS facilitates the conversion of the discrete architecture into a format amenable to gradient-based optimization. In our DAG-based search space, edges connect pairs of cells, and each edge carries a set of candidate operations. For an edge connecting cell $p$ to cell $q$ in layer $l$, the output is computed as follows:

\begin{align}
	\label{eq10}
	{\bar{o}}^{(p,q)}(X)=\sum_{o \in \mathcal{O}^{(l,p,q)}}\frac{\exp(\alpha_{o}^{(l,p,q)})}{\sum_{o' \in \mathcal{O}^{(l,p,q)}}\exp(\alpha_{o'}^{(l,p,q)})}\ o(X),
\end{align}
where $\alpha_{o}^{(l,p,q)} \in \mathbb{R}$ is a learnable scalar quantifying the contribution of candidate operation $o$ on edge $(p,q)$. Crucially, $p$ and $q$ index **cells in the DAG supernetwork** (e.g., the input cell, operation cells, and output cell), not nodes in the original input graph. In the first stage, each operation cell $i$ is directly connected from the input cell to the output cell, forming a star topology. Consequently, the general DARTS weight $\alpha_{o}^{(l,p,q)}$ reduces to the simpler layer-operation notation $\alpha_{S}^{(l,i)}$ used in Eq.~\eqref{XS1}. Similarly, in the second stage, it reduces to $\alpha_{G}^{(l,i)}$ in Eq.~\eqref{G}. The task of architecture search is thus simplified to learning the continuous variables $\alpha_{S}^{(l,i)}$ and $\alpha_{G}^{(l,i)}$, a problem that can be effectively tackled using DARTS.

\subsubsection{Discretization and Final Architecture}

Once the search phase is complete, the continuous architecture weights are converted into a discrete architecture before retraining. Following the standard DARTS protocol \cite{liu2018darts}, for each layer $l$, we select the operation with the largest weight and discard all others:

\begin{align}
	\label{discrete}
	i_{S}^{(l)} &= \arg\max_{i} \alpha_{S}^{(l,i)},\quad
	i_{G}^{(l)} = \arg\max_{i} \alpha_{G}^{(l,i)},
\end{align}
where $i_{S}^{(l)}$ denotes the selected node scoring strategy and $i_{G}^{(l)}$ denotes the selected graph neural network operation for the $l$-th layer. This yields a fixed, layer-specific architecture that is then retrained from scratch using the standard training procedure described in Section~4. By deriving a discrete architecture in this manner, we retain the flexibility of the continuous search while producing a compact and efficient model for deployment.

\subsection{Interpretability and Explainability}

In graph learning, interpretable models are those that can provide human-understandable mechanisms to make predictions \cite{yuan2022explainability}, while results by explainable models are understandable by post hoc explanation techniques \cite{xuanyuan2023global}. Our proposed approach enhances interpretability by learning a soft gating vector for each layer from seven different scoring perspectives, where each entry indicates how much a node contributes to subsequent GNN processing. By assigning low gate values to nodes that require less further processing, the proposed approach can effectively reduce over-smoothing and enhance discrimination power. For visualization purposes, we binarize the continuous gate values with a threshold of $0.5$ to highlight which nodes are strongly selected (gate $>0.5$, shown in red) versus weakly selected (gate $\le 0.5$, shown in green), as detailed in the experiments.

In short, the explicit soft gating mechanism provides valuable insights into improving neural network architectures, offering a controllable and effective information filtering mechanism for GNNs, thereby optimizing the learning and information propagation process across deep layers.

\section{Experiments}
\subsection{Experimental setting}

\textbf{Datasets and Tasks.} We evaluated our approach on two task categories—graph classification and node classification—utilizing eight different datasets: Cora, CiteSeer, PubMed, D\&D, PROTEINS, IMDB-MULTI, COX2, and MR. The Cora, CiteSeer, and PubMed datasets are publication citation networks where nodes represent documents and edges denote citations, adhering to the standard train/validation/test split proposed by Yang et al \cite{yang2016revisiting}. The D\&D and PROTEINS datasets encode protein structures, with nodes representing amino acids and edges formed if the distance between them is less than 6 angstroms. The IMDB-MULTI dataset focuses on movie collaborations, detailing actor/actress and genre information sourced from IMDB, where nodes correspond to actors/actresses and edges indicate their co-appearance in movies. The COX2 dataset includes 467 cyclooxygenase-2 (COX-2) inhibitors, where these compounds are classified based on their activity levels as either active or inactive against human recombinant enzymes in vitro. The MR dataset consists of movie reviews for binary sentiment classification, featuring 5331 positive and 5331 negative reviews, each represented as a single sentence. In these graphs, nodes denote unique words, and edges represent co-occurrences of word pairs within a fixed-size sliding window. Detailed information about the eight datasets can be found in Table \ref{tab:2}.

\begin{table*}[htbp]
	\caption{Statistics of the datasets used in the experiments. For node classification tasks, the terms "Nodes" and "Edges" correspond to the cumulative count of nodes and all edges within the datasets. Conversely, "Nodes" and "Edges" signify the mean number of nodes and edges across the graphs for graph classification tasks. }
	\label{tab:2}       
	\centering
	\begin{tabular}{ccccccc}
		\hline
		Task                                  & Dataset    & Graphs & Nodes & Edges & Features & Classes \\ \hline
		\multirow{3}{*}{Node Classification}  & Cora       & -      & 2708  & 10556 & 1433     & 7       \\
		& Citeseer   & -      & 3327  & 9104  & 3703     & 6       \\
		& PubMed     & -      & 19717 & 88648 & 500      & 3       \\ \hline
		\multirow{5}{*}{Graph Classification} & D\&D       & 1178   & 384.3 & 715.7 & 89       & 2       \\
		& PROTEINS  & 1113   & 39.1  & 72.8  & 3        & 2       \\
		& IMDB-MULTI & 1500   & 13    & 65.9  & 0        & 3       \\
		& COX2       & 467    & 41.2  & 43.5  & 3        & 2       \\
		& MR         & 10654  & 19.5  & 73.5  & 300      & 2       \\ \hline
	\end{tabular}
\end{table*}

\textbf{Compared Methods.} We conduct a comparative analysis of N-GNAS against manually crafted GNN architectures and existing NAS methods. The manually crafted baselines encompass: GCN \cite{kipf2016semi}, GIN \cite{xu2018powerful}, GraphSAGE \cite{hamilton2017inductive}, GAT \cite{velivckovic2017graph}, SGC \cite{wu2019simplifying}, alongside more powerful baselines such as PNA \cite{corso2020principal}, AP-GCN \cite{spinelli2021adaptive}, and DGCNN \cite{zhang2018end}. Our experiments utilize the widely recognized, open-source library PyTorch Geometric (PyG) \cite{fey1903fast}, which facilitates the implementation of diverse GNN models. We conduct training for each baseline five times across various datasets, using the optimal hyperparameters specified in the paper. We present the mean accuracy and variability measure to assess the models' evaluative performance. Furthermore, we compare with a suite of GNAS baselines, including (I) GraphNAS \cite{gao2021graph}, an RL-based method incorporating a parameter-sharing scheme. (II) SNAG \cite{zhao2020simplifying}, an RL-based GNAS method emphasizing the combination of nodes and layers. (III) SANE \cite{huan2021search}, a GD-based method that operates on a search space comprising the combination of nodes and layers along with skip-connections (IV) PAS \cite{wei2023neural}, a GD-based GNAS method approach that integrates Aggregation, Pooling, Readout, and Merge components. (V) Genetic-GNN \cite{shi2022genetic}, an EA-based GNAS method that dynamically adapts the interplay between the GNN model structure and hyperparameters through an alternating evolutionary process, aiming to find the optimal combination of GNN model structure and hyperparameters. (VI) KEGNAS \cite{knowledge-aware}, a knowledge‑aware EA-based GNAS method that transfers prior knowledge from NAS‑Bench‑Graph\cite{qin2022bench} via a knowledge model and a deep multi‑output Gaussian process (DMOGP) to warm‑start an evolutionary algorithm for optimizing accuracy and model size. (VII) DFG-NAS \cite{zhang2022deep}, an EA-based GNAS method concentrating on macro-architecture design, involves integrating and organizing propagation and transformation operations into GNNs to optimize GNN architectures. (VIII) ExGNAS\cite{exgnas}, an explainable Graph NAS method based on Monte‑Carlo tree search that improves interpretability via a simple search space and decision‑traceable architecture selection.

\textbf{Searching setting.} In our implementation of N-GNAS, we delineate a two-stage search space within each layer, comprising seven and five operational cells, respectively, denoted by $n_c=7,5$. To ensure experimental consistency, the depth of our model is consistently set to two layers. We allocate half of the available training data for the search process for validation purposes. We adopt DARTS as our search strategy, where the operation weights, architecture parameters, and the corresponding optimizer are configured similarly to previous works \cite{huan2021search, zhao2020simplifying}. Furthermore, we initiate the architecture search with distinct random seeds on eight diverse datasets, each employing a 2-layer GNN model throughout 100 epochs. The batch size is standardized to 64, and the node feature dimensionality is upscaled to 64. Through this sequence of eight architectural searches, we successfully identify the premier architectural configuration on most datasets, with the top-1 architecture prevailing on seven datasets and the top-2 on one remaining dataset.

\textbf{Training setting.} In the training phase, we follow the configurations employed in earlier studies \cite{cai2021rethinking}, which encompass data splitting, optimizer selection, feature dimension adjustment, batch normalization, and the inclusion of residual connections. In particular, we employed the Adam optimizer with an initial learning rate 1e-5 for different datasets. To alleviate overfitting, a dropout rate of 0.2 was used. We conducted five repetitions of the top-1 architecture identified on the validation dataset, each for 400 epochs, using a batch size of 64. Ultimately, We reported the average performance and standard deviation of the top-1 architecture on the validation dataset. All experiments used a single NVIDIA GeForce RTX 3090 GPU (with 24GB memory, CUDA version 11.2).

\subsection{Results and Analysis}

\begin{figure}
	\centering
	\includegraphics[width=86mm]{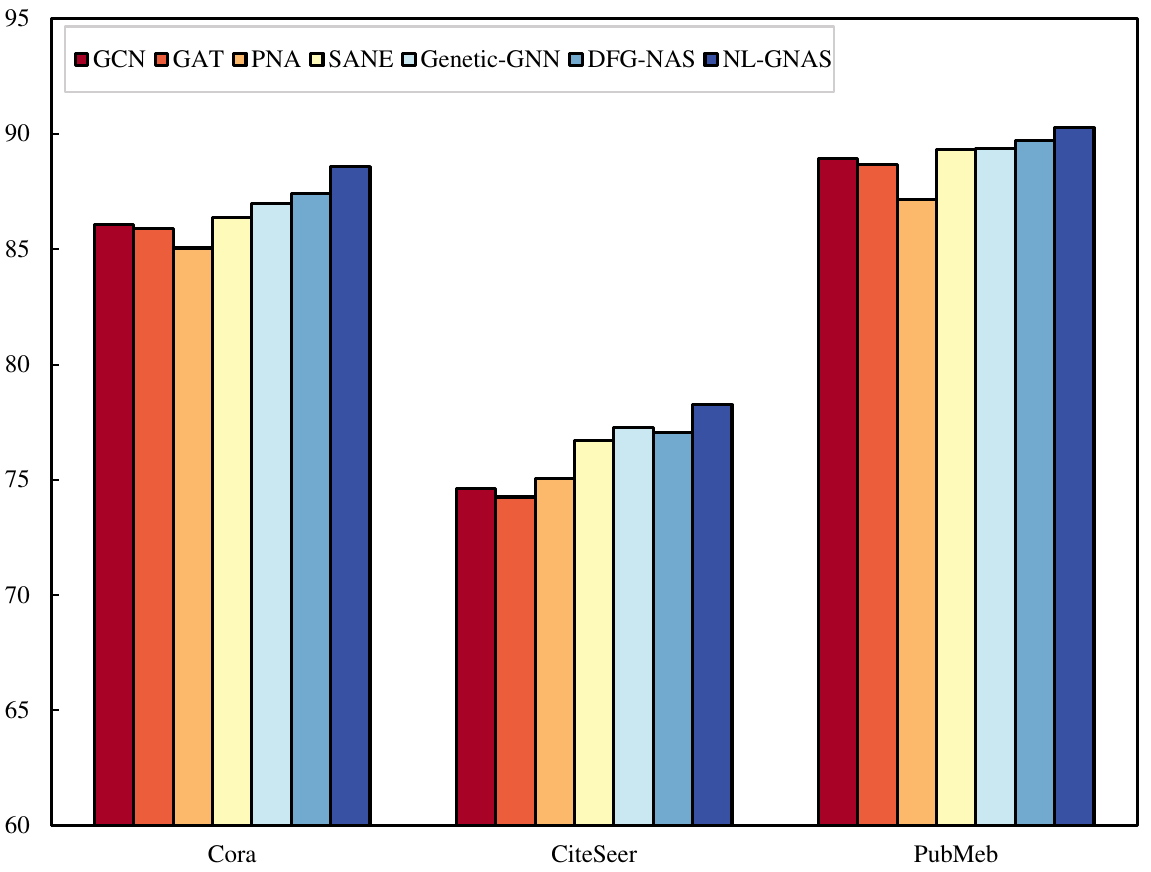}
	\caption{\textbf{Results of various methods on the Cora, CiteSeer, and PubMed datasets.} Different colors represent different methods. It can be found that our work achieves the best results on both datasets}
	\label{img:result}
\end{figure}

\textbf{Results on node classification task.} Cora originates from machine learning papers across various disciplines, CiteSeer is sourced from academic papers in computer science, and PubMed encompasses research content from multiple biomedical domains. These datasets are commonly employed for node classification tasks. The main reason is that current manually crafted and NAS methods in architecture design consider using the same number of graph neural networks for each node, leading to over-smoothing. Table \ref{tab:node-task} and Figure \ref{img:result} present the results of the node classification task on the Cora, CiteSeer, and PubMed datasets. Our node classification performance surpasses current models, showing a 1.3\% increase in test accuracy, highlighting our approach's efficacy.

\begin{table*}[ht]
	\centering
	\caption{Node classification task accuracy on the Cora, CiteSeer, and PubMed datasets. The top three are emphasized by \textcolor{red}{\textbf{first}}, \textcolor{orange}{\textbf{second}}, and \textbf{third.}}
	\label{tab:node-task}
	\begin{tabular}{c|c|ccc}
		\hline
		& Methods     & Cora       & CiteSeer   & PubMed     \\ \hline
		& GCN \cite{kipf2016semi}         & 86.09±0.50 & 74.64±0.20 & 88.96±0.29 \\
		& GIN \cite{xu2018powerful}         & 85.68±0.61 & 73.40±0.14 & 88.23±0.28 \\
		& GraphSAGE \cite{hamilton2017inductive}   & 85.66±0.52 & 74.59±0.63 & 89.21±0.29 \\
		& GAT \cite{velivckovic2017graph}         & 85.92±0.72 & 74.26±0.13 & 88.67±0.19 \\
		& SGC \cite{wu2019simplifying}         & 85.31±0.86 & 72.94±0.98 & 88.40±0.25 \\
		& PNA \cite{corso2020principal}         & 85.06±0.72 & 75.06±0.61 & 87.18±0.30 \\
		\multirow{-7}{*}{manually crafted}             & AP-GCN \cite{spinelli2021adaptive}      & 86.75±0.29 & 75.64±0.75 & \textcolor{orange}{\textbf{89.79±0.26}} \\ \hline
		{\color[HTML]{643403} }                      & GraphNAS \cite{gao2021graph}    & 86.69±0.71 & 76.32±0.61 & 87.93±0.16 \\
		{\color[HTML]{643403} }                      & SNAG \cite{zhao2020simplifying}        & 84.99±1.04 & 75.94±1.27 & 88.44±0.25 \\
		{\color[HTML]{643403} }                      & SANE \cite{huan2021search}        & 86.40±0.38 & 76.72±0.98 & 89.34±0.31 \\
		{\color[HTML]{643403} }                      & Genetic-GNN \cite{shi2022genetic} & {86.99±0.60} & \textcolor{orange}{\textbf{77.29±0.72}} & 89.37±0.26 \\
		& KEGNAS \cite{knowledge-aware} & 82.44±0.23 & 71.05±0.22 & 78.24±0.36 \\
		{\color[HTML]{643403} }                      & DFG-NAS \cite{zhang2022deep}     & {\textbf{87.42±0.42}} & \textbf{77.04±0.68} & \textbf{89.73±0.20} \\ 
		& ExGNAS \cite{exgnas} & \textcolor{orange}{\textbf{88.50±0.01}}  & 74.7±0.02 & 89.3±0.10 \\ \cline{2-5}
		\multirow{-6}{*}{{\color[HTML]{643403} NAS}} & N-GNAS       & \textcolor{red}{\textbf{88.59±0.63}} & \textcolor{red}{\textbf{78.26±0.42}} & \textcolor{red}{\textbf{90.27±0.48}} \\ \hline
	\end{tabular}
\end{table*}

\textbf{Results on graph classification task.} Our model's performance in graph classification tasks was assessed through experiments conducted on the D\&D, PROTEINS, IMDB-MULTI, COX2, and MR datasets. The results in Table \ref{tab:graph-task} show that our model achieves excellent results in this domain, similar to the node classification tasks. Compared to manually crafted and NAS methods, our model outperforms the baselines by 0.7\% on the D\&D dataset and by 0.6\% on the PROTEINS dataset.

\begin{table*}[ht]
	\centering
	\caption{Graph classification task accuracy on the D\&D, PROTEINS, IMDB-MULTI, COX2, and MR datasets. The top three are emphasized by \textcolor{red}{\textbf{first}}, \textcolor{orange}{\textbf{second}}, and \textbf{third.}}
	\label{tab:graph-task}
	\begin{tabular}{c|c|ccccc}
		\hline
		& Methods   & D\&D       & PROTEINS   & IMDB-MULTI & COX2       & MR         \\ \hline
		& GCN \cite{kipf2016semi}       & \textbf{76.98±4.43} & 72.94±1.82 & \textcolor{orange}{\textbf{50.25±3.42}} & 78.68±1.92 & 75.62±0.85 \\
		& GIN \cite{xu2018powerful}       & 73.95±2.98 & 73.68±2.78 & 50.04±2.75 & 80.52±3.41 & 76.05±0.74 \\
		& GraphSAGE \cite{hamilton2017inductive} & 76.78±4.06 & 72.58±2.43 & 49.62±4.58 & 79.63±2.63 & 76.85±0.63 \\
		& GAT \cite{velivckovic2017graph}       & 75.14±2.84 & \textbf{74.29±1.69} & 49.85±3.65 & \textbf{81.16±3.64} & 76.92±1.02 \\
		\multirow{-5}{*}{manually crafted}             & DGCNN \cite{zhang2018end}     & 76.66±4.03 & 73.28±3.16 & 49.63±3.73 & 80.16±3.23 & \textbf{77.06±0.68} \\ \hline
		{\color[HTML]{643403} }                      & GraphNAS \cite{gao2021graph}  & 73.56±2.47 & 73.12±4.27 & 47.23±4.59 & 78.91±2.37 & 76.37±2.17 \\
		{\color[HTML]{643403} }                      & SNAG \cite{zhao2020simplifying}      & 74.46±2.69 & 71.98±3.42 & 49.38±1.79 & 79.32±2.74 & 76.23±2.61 \\
		{\color[HTML]{643403} }                      & SANE \cite{huan2021search}      & 73.62±3.82 & 72.23±3.86 & 49.92±3.71 & 80.04±2.62 & \textcolor{orange}{\textbf{77.27±2.12}} \\
		{\color[HTML]{643403} }                      & PAS \cite{wei2023neural}       & \textcolor{orange}{\textbf{77.46±3.68}} & \textcolor{orange}{\textbf{74.85±2.93}} & \textbf{50.07±2.94} & \textcolor{red}{\textbf{82.16±3.95}} & 76.72±2.29 \\ \cline{2-7} 
		\multirow{-5}{*}{{\color[HTML]{643403} NAS}} & N-GNAS     & \textcolor{red}{\textbf{78.03±3.64}} & \textcolor{red}{\textbf{75.30±3.27}} & \textcolor{red}{\textbf{51.62±3.51}}        &      \textcolor{orange}{\textbf{81.29±4.17}}      &     \textcolor{red}{\textbf{78.27±1.95}}      \\ \hline
	\end{tabular}
\end{table*}

\begin{figure}
	\centering
	\includegraphics[width=86mm]{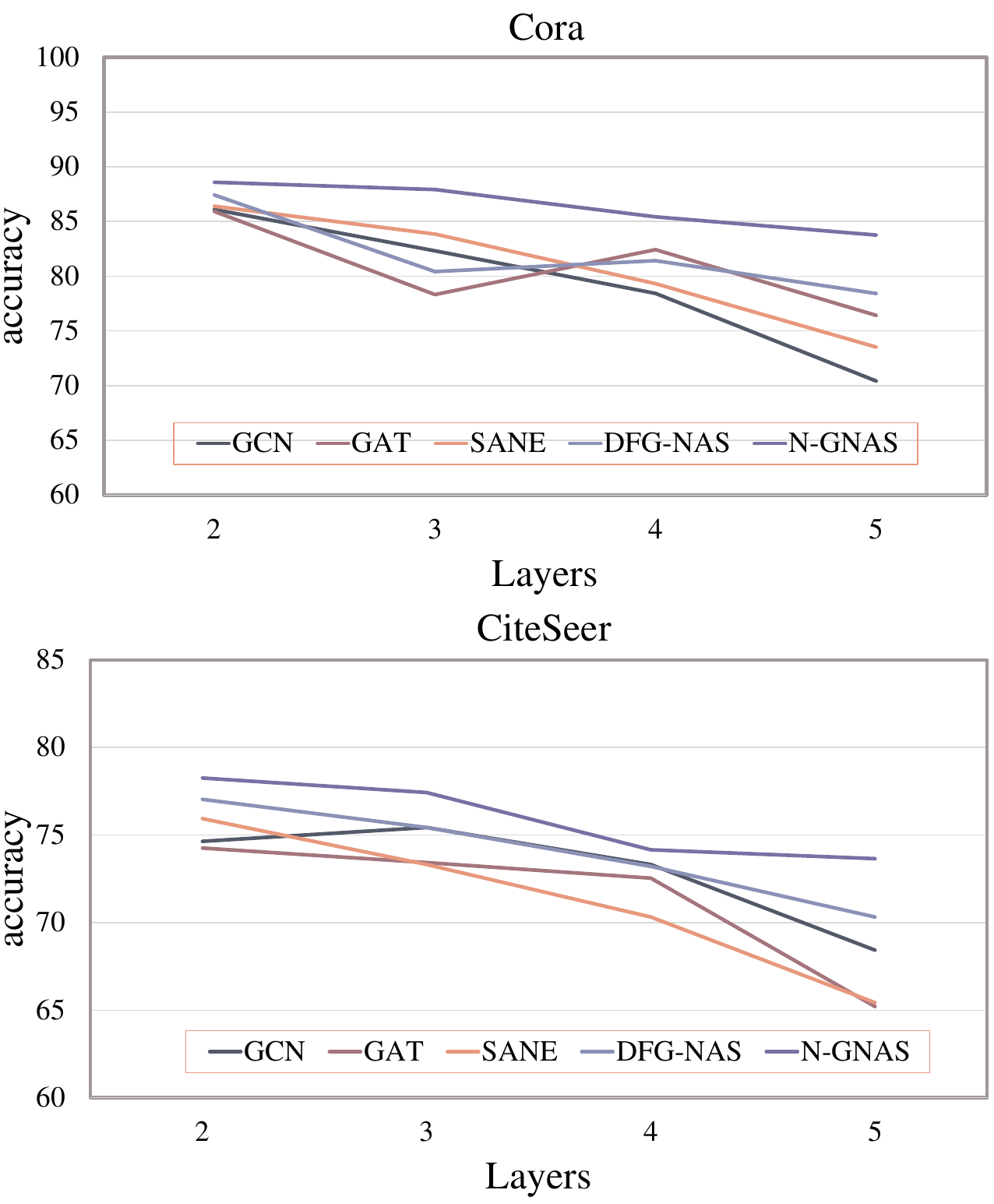}
	\caption{\textbf{Performance of the model with different numbers of layers.} Different colors represent different methods. It can be found that our work achieves the best results on both datasets}
	\label{different layers}
\end{figure}
\begin{figure}
	\centering
	\subfloat[the first layer]
	{
		\includegraphics[width=43mm]{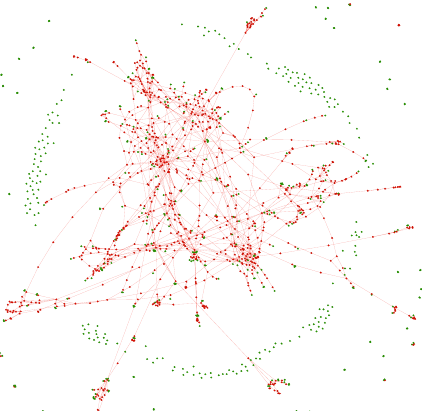}
		\label{local structural}
	}
	\subfloat[the second layer]
	{
		\includegraphics[width=43mm]{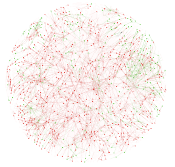}
		\label{global position}
	}
	\caption{\textbf{Node selection results obtained by the first and the second layer of graph neural networks optimized with N-GNAS.} Red nodes (gate value $>0.5$) are strongly selected for GNN processing, while green nodes (gate value $\le 0.5$) mostly bypass GNN operations via residual connections.}
	\label{img:explainability}

\end{figure}

\textbf{Discussion of different layers.} As illustrated in Figure \ref{different layers}, we investigate the performance of our method on the Cora and PubMed datasets across various layers. We compared N-GNAS with several baselines, including GCN, GAT, SANE, and DFG-NAS, in terms of accuracy for different numbers of layers. As depicted in Figure \ref{different layers}, we conclude that GNAS's performance remains stable as the layers' depth increases, outperforming other baselines. However, the performance of the baselines relative to N-GNAS is lower, and it decreases as the depth increases, indicating the occurrence of over-smoothing issues in these baselines. Therefore, N-GNAS can simultaneously consider model performance and alleviate over-smoothing problems.

\textbf{Explainability of the search results.} We conducted comprehensive experiments to explain the results on the Cora dataset, as illustrated in Figure \ref{img:explainability}. The first layer of the obtained network determined by N-GNAS uses the fifth metric, the local structural feature scoring, as the node selection metric, and thus, the selected nodes in red are typical with larger degrees, as shown in Figure \ref{local structural}. In the second layer, the fourth metric, the global position scoring, is used as the node selection metric, indicating that more central nodes with shorter overall distances to others are more likely to be selected, as shown in Figure \ref{global position}. These results demonstrated the explainability. Moreover, the prediction quality benefits directly from this mechanism: as shown in Table \ref{tab:explainability}, the accuracy gap between N-GNAS and its variant without node selection widens from 1.58\% (1 layer) to 4.19\% (5 layers), confirming that the learned gating structure becomes increasingly critical for maintaining discriminative predictions in deeper models.

The explainability can also be perceived from the ablation study, as shown in Table \ref{tab:explainability}. Over-smoothing can significantly deteriorate model effectiveness in traditional GNNs as the number of layers increases. Nevertheless, the N-GNAS framework mitigates the negative impact of increased layer depth on performance, which indicates that it reduces the over-smoothing by only using selected nodes for further processing.

\begin{table}[htp]
	\centering
	\caption{Performance of N-GNAS with varying layers on the Cora dataset. We present the test accuracy results for N-GNAS without Node Selection Operations and for N-GNAS with these operations.}
	\label{tab:explainability}
	\begin{tabular}{ccc}
		\hline
		\multicolumn{3}{c}{Cora}        \\ \hline
		Layer & Stage II   & N-GNAS      \\ \hline
		1     & 76.93±0.49 & 78.51±1.13 \\
		2     & 86.36±1.12 & 88.59±0.63 \\
		3     & 84.43±0.71 & 87.92±0.68 \\
		4     & 82.71±0.83 & 85.43±0.66 \\
		5     & 79.57±0.55 & 83.76±0.47 \\ \hline
	\end{tabular}
\end{table}

\textbf{Ablation study.} An ablation study was carried out to showcase the efficacy of our custom-designed search space and loss function. Table \ref{tablation study} compares N-GNAS against two baseline models: (1) Random: randomly selecting one architecture without neural architecture search, and (2) $L_{ce}$: using only the cross-entropy loss $L_{ce}$ in the loss function while turning off $L_{cl}$. For models that did not utilize our search space and instead employed a randomly chosen fixed architecture, the accuracy decreased by 1.8-4.6\%. This decrease is attributed to the inability of the model to assign appropriate graph neural network counts to different nodes without using the proper Node-wise optimal convolution selection operation and Graph convolution, leading to performance degradation. For models that did not employ $L_{cl}$ and only used $L_{ce}$, the accuracy decreased by 1.8-2.5\%. This decrease is because not using $L_{cl}$ only considers the model's performance while ignoring the over-smoothing issue, posing a risk of over-smoothing. These ablation studies indicate the effectiveness of our designed search space and loss function.

\begin{table}[ht]
	\centering
	\caption{Ablation study on search space and loss function.}
	\label{tablation study}
	\begin{tabular}{cccc}
		\hline
		Methods   & Cora  & CiteSeer & PubMed \\ \hline
		Random    & 85.74 & 74.63    & 88.62  \\
		$L_{ce}$ & 86.94 & 76.52    & 87.98  \\ \hline
		N-GNAS     & 88.59 & 78.26    & 90.27  \\ \hline
	\end{tabular}
\end{table}

\section{Conclusion}
This work proposes an innovative N-GNAS model for graph neural architecture search. This framework is distinguished by its adaptability and generality, featuring a dynamic search space for the optimal count of graph neural networks. It facilitates the discovery of superior GNN configurations by integrating node-wise optimal convolution selection and graph neural network operation. A novel loss function has been engineered to balance model performance challenges and the over-smoothing phenomenon. It demonstrates strong competitiveness against manually crafted GNNs across various tasks. Furthermore, a two-phase search approach is presented to enhance search efficiency. Broad experimental results spanning multiple graph learning tasks demonstrate the effectiveness of our method, with graphical illustrations of the outcomes reinforcing widely accepted insights.






\printcredits

\bibliographystyle{model1-num-names}

\bibliography{mybib}



\end{document}